\documentclass[conference]{IEEEtran}
\IEEEoverridecommandlockouts

\usepackage{cite}
\usepackage{amsmath,amssymb,amsfonts}
\usepackage{algorithmic}
\usepackage{graphicx}
\usepackage{textcomp}
\usepackage{xcolor}
\usepackage{subcaption}
\usepackage{enumitem}

\def\BibTeX{{\rm B\kern-.05em{\sc i\kern-.025em b}\kern-.08em
    T\kern-.1667em\lower.7ex\hbox{E}\kern-.125emX}}
\begin{document}

\title{Application of frozen large-scale models to multimodal task-oriented dialogue\\
}

\author{\IEEEauthorblockN{1\textsuperscript{st} Tatsuki Kawamoto}
\IEEEauthorblockA{\textit{Meiji University}\\
Kanagawa, Japan \\
tatsuki.00.0306@gmail.com}
\and
\IEEEauthorblockN{2\textsuperscript{nd} Takuma Suzuki}
\IEEEauthorblockA{\textit{Meiji University}\\
Kanagawa, Japan \\
stakuma9912@gmail.com}
\and
\IEEEauthorblockN{3\textsuperscript{rd} Ko Miyama}
\IEEEauthorblockA{\textit{Meiji University}\\
Kanagawa, Japan \\
chinpanzi914@gmail.com}
\and
\IEEEauthorblockN{4\textsuperscript{th} Takumi Meguro}
\IEEEauthorblockA{\textit{Meiji University}\\
Kanagawa, Japan \\
takumimeguro1321@gmail.com}
\and
\IEEEauthorblockN{5\textsuperscript{th} Tomohiro Takagi}
\IEEEauthorblockA{\textit{Meiji University}\\
Kanagawa, Japan \\
takagit@gmail.com}
}

\maketitle

\begin{abstract}

In this study, we use the existing Large Language Models ENnhanced to See Framework (LENS Framework) to test the feasibility of multimodal task-oriented dialogues. The LENS Framework has been proposed as a method to solve computer vision tasks without additional training and with fixed parameters of pre-trained models. We used the Multimodal Dialogs (MMD) dataset, a multimodal task-oriented dialogue benchmark dataset from the fashion field, and for the evaluation, we used the ChatGPT-based G-EVAL, which only accepts textual modalities, with arrangements to handle multimodal data. Compared to Transformer-based models in previous studies, our method demonstrated an absolute lift of 10.8\% in fluency, 8.8\% in usefulness, and 5.2\% in relevance and coherence. The results show that using large-scale models with fixed parameters rather than using models trained on a dataset from scratch improves performance in multimodal task-oriented dialogues. At the same time, we show that Large Language Models (LLMs) are effective for multimodal task-oriented dialogues. This is expected to lead to efficient applications to existing systems.
\end{abstract}

\begin{IEEEkeywords}
Multimodal task-oriented dialogues, Large Language Models
\end{IEEEkeywords}

\begin{figure}[htbp]
\centerline{\includegraphics[width=0.7\linewidth]{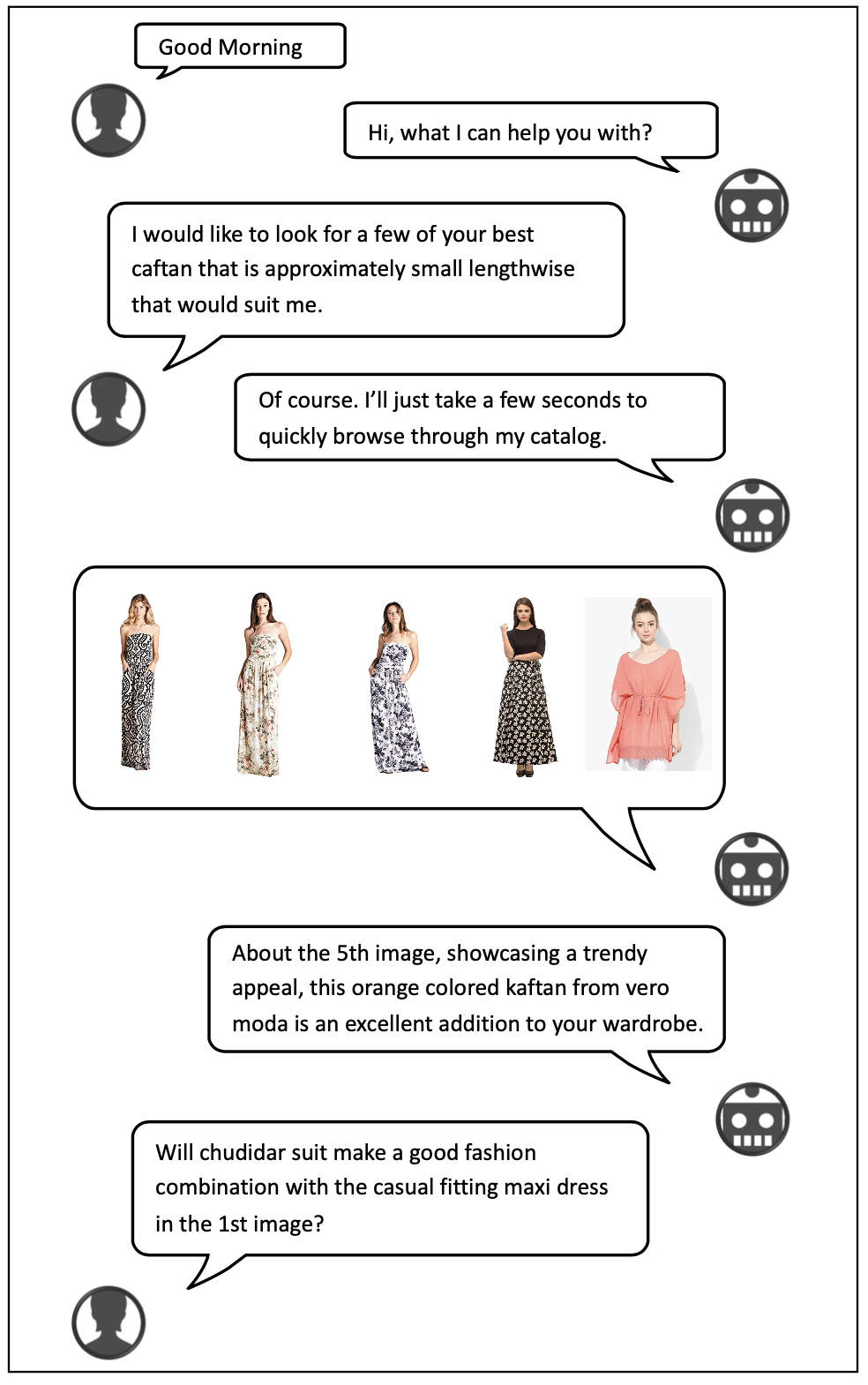}}
\caption{Example of dialogue in MMD}
\label{fig}
\end{figure}

\section{Introduction}
Dialogue systems have been the central subject of much research in natural language processing. These systems are divided into two main categories, open-domain dialogues and task-oriented dialogues. Open-domain dialogues are aimed at free dialogues on a wide range of topics, while task-oriented dialogues focus on dialogues to achieve a specific goal or task. Task-oriented dialogue research has evolved toward achieving more sophisticated and flexible dialogue. Within this context, multimodal task-oriented dialogues have emerged and have attracted much attention.\\
\indent  Multimodal task-oriented dialogues utilize not only textual information but also other modal (e.g., image and audio) information to complement and improve the accuracy of information. Nuances and information that cannot be conveyed by textual information alone can be complemented by other modalities. However, the realization of multimodal task-oriented dialogue involves various technical hurdles, such as the association of information between different modalities and the difficulty of learning, which requires a large data set. While many prior studies \cite{b1,b2,b3,b4,b5,b6,b7,b9,b10,b11} have been conducted to solve these problems, there have not been recent studies using large-scale models released by OpenAI, Meta, Google, and other organizations. The quantity and quality of knowledge, comprehension, and reasoning power of recent large-scale models have not been applied despite the fact that they are distinct from previous pre-trained models. We therefore adopted the Large Language Models ENnhanced to See Framework (LENS Framework) \cite{b21}, which uses fixed pre-trained models (also using a Large Language Model (LLM)) for computer vision problems without additional model training. The LENS Framework uses an LLM as inference modules. Currently, Large Multimodal Models (LMMs) exist in addition to LLMs. However, because LLMs are being more actively developed, we chose the LENS Framework because we believe it can be useful for real-world applications. BLIP2 \cite{b22} is used as the caption generation model and Llama2 \cite{b23} as the LLM. For simplicity of experimentation, we did not generate tags and attributes using CLIP \cite{b24}, which was used in the original paper.\\
\indent We used the Multimodal Dialogs (MMD) dataset \cite{b1}, a multimodal task-oriented dialogue dataset specialized for the fashion field, and compared its performance with an existing Transformer-based model \cite{b6}. We evaluated the performance and effectiveness of the framework by using G-EVAL \cite{b25}, which uses ChatGPT \cite{b17}, which only accepts text-modal data, with arrangements to handle multimodal data. Through such an evaluation, we provide insight into how to effectively implement multimodal task-oriented dialogues in real-world applications.\\
\indent The main contributions of our study are as follows.
 \begin{itemize}
  \item We proved that using large-scale models with fixed parameters is more suitable for multimodal task-oriented dialogues on various perspectives than using models trained from scratch on large datasets.
  \item We proved that multimodal task-oriented dialogues are feasible using LLMs that handle only text.
  \item We also extended G-EVAL, which uses ChatGPT as an evaluator, so that it can evaluate multimodal task-oriented dialogues. By applying our method, G-EVAL could be applied not only to multimodal task-oriented dialogues, but also to general multimodal tasks with text output.
 \end{itemize}

\section{Related works}
\subsection{Multimodal task-oriented dialogue}
There is a growing demand for multimodal chatbots capable of conversing via images, text, and audio in several domains, including retail, travel, and entertainment. However, the lack of large open datasets has hindered the progress of deep learning research. Against this backdrop, \cite{b1} constructed MMD dataset for the fashion domain consisting of over 150K conversational sessions. They also presented the basic Multimodal HieRarchical Encoder-Decoder model (MHRED) in the same paper, along with the dataset. \cite{b2} presented the Multimodal diAloG system with adaptIve deCoders (MAGIC), which can dynamically generate general, knowledge-aware, and multimodal responses based on various user intentions. In addition to the work we have just briefly described, many other studies have been conducted \cite{b3,b4,b5}. However, all existing methods are MHRED-based and cannot learn dependencies between multimodal semantic elements. Therefore, \cite{b6} proposed a Transformer-based method, Multi-modal diAlogue system with semanTic Elements (MATE), which can effectively handle dependencies between multimodal semantic elements. To address the problem in previous studies using MHRED that the interaction between textual and visual features is not sufficiently fine-grained and that the context representation cannot provide a complete representation of the context, \cite{b7} proposed a non-hierarchical attention network with modality dropout. This architecture also takes a structure similar to Transformer \cite{b8}. More recent work has also used pre-trained models, such as BART \cite{b12} and DialoGPT \cite{b13}, as seen in \cite{b9,b10,b11}. However, there has not been research on multimodal task-oriented dialogues using even larger models that have been published by OpenAI, Meta, Google, and other organizations. We focused our attention on this and conducted an evaluation of methods using large-scale models. In addition, Vision and Language tasks such as Image Captioning \cite{b14}, Visual Question Answering \cite{b15}, and Visual Dialog \cite{b16} are also relevant to our work. However, multimodal dialogue systems focus on multi-turn and multimodal interaction between user and assistant and are not fundamentally limited to a single image.
\subsection{Using LLMs in task-oriented dialogue}
Research has been conducted on using ChatGPT \cite{b17} and other LLMs to perform task-oriented dialogues with no or only a small amount of training, rather than training a model from scratch. For example, \cite{b18} conducted a study using an instruct-tuned LLM to evaluate its ability to complete multi-turn tasks and interact with external databases using task-oriented dialog benchmarks. Also, \cite{b19} introduced a framework, Directional Stimulus Prompting, which provides guidance for downstream tasks for a parameter-fixed LLM by tuning the language model, and evaluated it using a task-oriented dialog benchmark. In \cite{b20}, Schema-Guided Prompting for building Task-Oriented Dialog systems effortlessly based on LLMs (SGP-TOD) was proposed and evaluated using task-oriented dialogue benchmarks, which instructed a fixed LLM to generate appropriate responses to new tasks, avoiding the need for training data by using a task schema. However, to the best of our knowledge, there are no studies that use and evaluate large-scale models, including LLMs, for multimodal task-oriented dialogues.
\subsection{Pre-trained models in the Multimodal field}
In the multimodal field, there are pre-trained models such as CLIP \cite{b24}, which embeds images and text in the same space, and BLIP \cite{b26}, which supports both visual language understanding and visual language generation. In addition, a number of models have been released that are tuned to LLMs, such as LLaVA \cite{b27}, miniGPT-4 \cite{b28}, and OpenFlamingo \cite{b29}. These models require additional multimodal training, which is expensive in terms of data collection and training costs. In response, the LENS Framework was recently proposed to solve multimodal tasks using pre-trained models with fixed parameters (using an LLM as inference modules) without the need for additional multimodal training. The development of LLMs has also been remarkable compared to the large-scale multimodal models enumerated in this section.
 
\begin{figure}[htbp]
\centerline{\includegraphics[width=\linewidth]{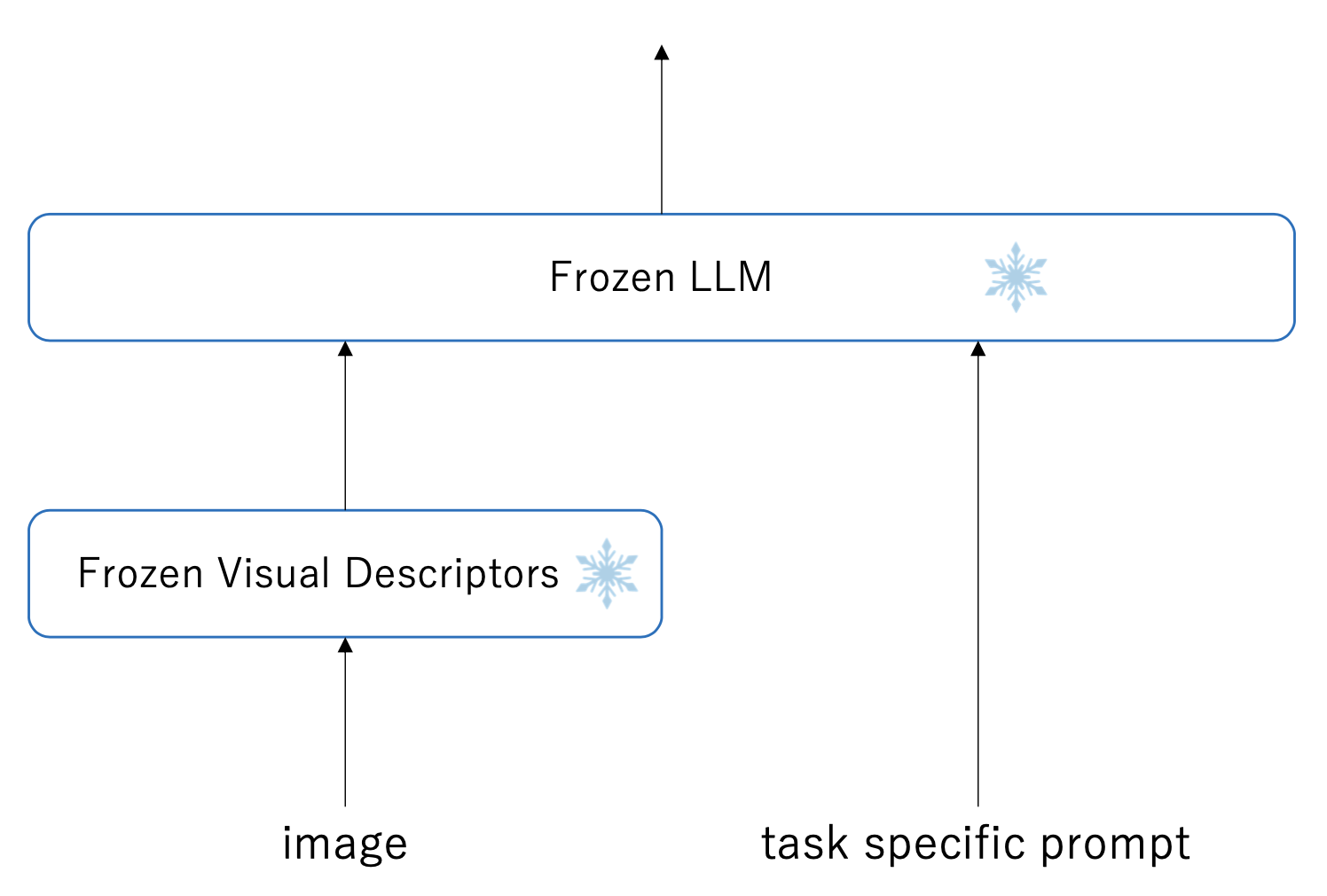}}
\caption{LENS Framework}
\label{fig}
\end{figure}

\begin{figure}[htbp]
\centerline{\includegraphics[width=0.9\linewidth]{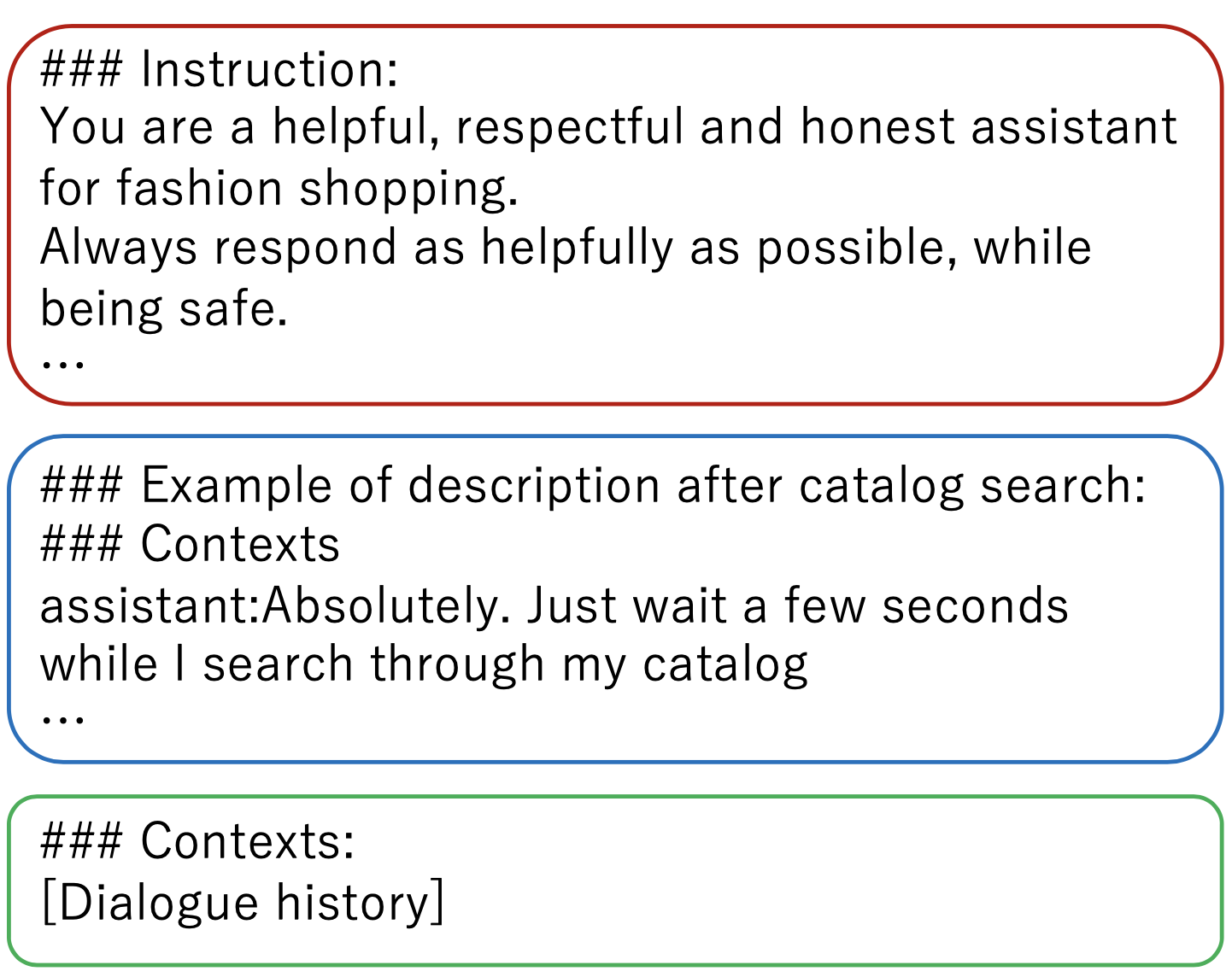}}
\caption{Prompt overview. The red box contains descriptions and instructions for tasks and datasets, the blue box contains dialogue examples, and the green box contains dialogue contexts.}
\label{fig}
\end{figure}
 
\section{Method}
We used the LENS Framework shown in Figure 2. The reasons for the choice of the framework have been discussed in the previous chapter. We first describe this framework, followed by a description of the prompts used in this study.

\subsection{LENS Framework}
In general, when working on Vision and Language tasks, including Image Captioning \cite{b14} and Visual Question Answering \cite{b15}, multimodal training is performed using image and language data. However, in this framework, by first generating tags, attributes, and captions using CLIP of a vision encoder and BLIP of an image captioning model, and then using the textual information and task instructions of the generated images as prompts to the inference module, an LLM, multimodal tasks can be performed without multimodal training.

\subsection{Prompt}
An overview of the prompts we used is shown in Figure 3. As mentioned previously, apart from the textual information in the generated images, task instructions are required for the prompt. We assumed the MMD dataset to be a real-world fashion shopping scenario between a user and an assistant, and described in the prompt what the dataset was and how it should behave for an equal comparison with previous studies. This approach allows for a fair comparison with the existing methods.\\
\indent The MMD dataset includes searching the catalog, presenting the item to the user, and then providing useful information to the user about the presented item. Since it was difficult to make an LLM behave in zero-shot mode, we used only one example obtained from the training data and constructed a prompt in a one-shot setting. The actual prompts used are shown in the Appendix.
\cite{b7}.

\section{Experiments}
\subsection{Dataset}
We use the MMD dataset, a multimodal task-oriented dialogue dataset in the fashion domain. The dialogue takes place between two parties, the user and the assistant. The user states a request, and the assistant introduces various products step by step until they either sign a purchase contract or quit shopping. The dialogue seamlessly incorporates multimodal data from two domains: image and text. The dialogue also incorporates a wealth of fashion domain-specific knowledge: over one million fashion products and their semi/unstructured information from the online retail sites Amazon, Jabong, and Abof. Based on this dataset, \cite{b1} proposed a text response generation task, an image response task, an employing domain-knowledge, and a user modeling. Here, we briefly introduce the main tasks, the text response generation task and the image response task. The former task is to generate the next text response given a k-turn context. The latter task is to output the most relevant image based on its relevance to the given context. Note that the image response task is further divided into two parts: one task is to retrieve images from the database, and the other is to generate images. In our study, we performed the text response generation task.

\subsection{Experiment Setup}
The LENS Framework uses CLIP and BLIP to perform various multimodal tasks. In our study, for simplicity of experimentation, we use a caption generation model to generate a single caption for each image. As a caption generation model, we use BLIP2 (blip2-flan-t5-xl), which is more accurate than BLIP. Llama2-Chat (13B) is used as the LLM. The models were chosen due to the constraints of the experimental environment. Following previous studies \cite{b8, b27, b29}, we use the two utterances before the response as the context.

\subsection{Compared Methods}
To demonstrate the effectiveness of the LENS Framework with an LLM as its main architecture used in this paper, we compared it with the following methods.\\
\indent MATE \cite{b6} : A Transformer-based multimodal task-oriented dialogue system. This method solves the problems of previous studies in that it ignores the contextual dependencies between words and images, and in that it only considers images from the current turn and excludes images from previous turns and their ordinal information when integrating visual content. We chose to use this study as a comparative method because it is based on Transformer, the architecture on which recent methods are based. Also, due to reproducibility and time resource constraints, we use this method whose code is publicly available on github.\\
\indent The LENS Framework used in our study does not utilize the knowledge base of the MMD dataset. The MATE also trains and uses a model using a first-stage decoder that does not use the knowledge base.
\subsection{Evaluation Metrics}
As indicated by \cite{b25}, conventional quantitative evaluation indices such as BLEU \cite{b30} have a low correlation with human evaluation. Therefore, we did not use those evaluation indices but instead conducted a quantitative evaluation based on perspectives such as fluency and relevance, which are used in human evaluation (the evaluators are described in the next subsection).\\
\indent We randomly selected 50 examples from the test data for evaluation. The context of each example was input into LENS Framework and MATE, a comparison method, to generate text utterances. We then evaluated each utterance according to the following indices.
\begin{itemize}
	\item Fluency : Is the response smooth and grammatically correct?
	\item Relevance : Is the response pertinent to the question or context provided?
	\item Coherence : Does the response logically follow the question and stay on topic?
	\item Usefulness : Does the response provide value to the user?
 \end{itemize}

\begin{table*}[htbp]
\caption{Verifying ChatGPT's ability to understand in-context image captions and fashion domain knowledge}
\begin{center}
\begin{tabular}{|c|c|c|}
\hline
& \textbf{\textit{Can image captions be recognized accurately? (\%)}}& \textbf{\textit{Can ChatGPT accurately explain fashion terms? (\%)}}\\
\hline
\textbf{ChatGPT(GPT-4)}&\textbf{100.0}&\textbf{100.0}\\
\hline
\end{tabular}
\label{tab1}
\end{center}
\end{table*}

\subsection{Evaluation Methods}
We used the G-EVAL framework, which uses ChatGPT (GPT-4) as the evaluator. This is an evaluation framework that uses GPT-3.5 (text-davinci-003) or GPT-4 as the evaluator, rather than metrics such as BLEU or NIST or human evaluation. \cite{b25} showed that it outperforms state-of-the-art evaluators and achieves higher human correspondence in the summarization and text generation tasks. We chose this framework because of its accuracy, high experimental reproducibility, and rich domain knowledge. We cannot say that \cite{b25} can be used as an evaluator in studies on multimodal task-oriented dialogues, as it only proves that text generation tasks can be evaluated. Therefore, a simple validation using the MMD dataset verified that G-EVAL is effective in the current experimental setting (Table 1). Specifically, we tested for the ability to understand the in-context image captions and knowledge of the fashion domain in the MMD data. We randomly selected 10 data perspectives from the test data and verified the former with the task of having ChatGPT extract captions without over- or under-extracting them, and the latter with the task of having ChatGPT explain the fashion domain terms in the context. As a result, we were able to determine that both tasks were performed appropriately in all 10 cases. Based on these results, we concluded that G-EVAL works effectively in this experimental setting and conducted the present study.\\
\indent We used a 5-point scale from 1 to 5, with 1 being the worst and 5 being the best. The rating was set to be better than 3 (e.g., a response was considered fluent if rated higher than 3).\\
\indent Since the performance was evaluated using perspectives other than those validated in the G-EVAL paper, we created our own prompts. The prompts used in this study are listed in the Appendix.

\begin{figure*}[htbp]
\centerline{\includegraphics[width=\linewidth]{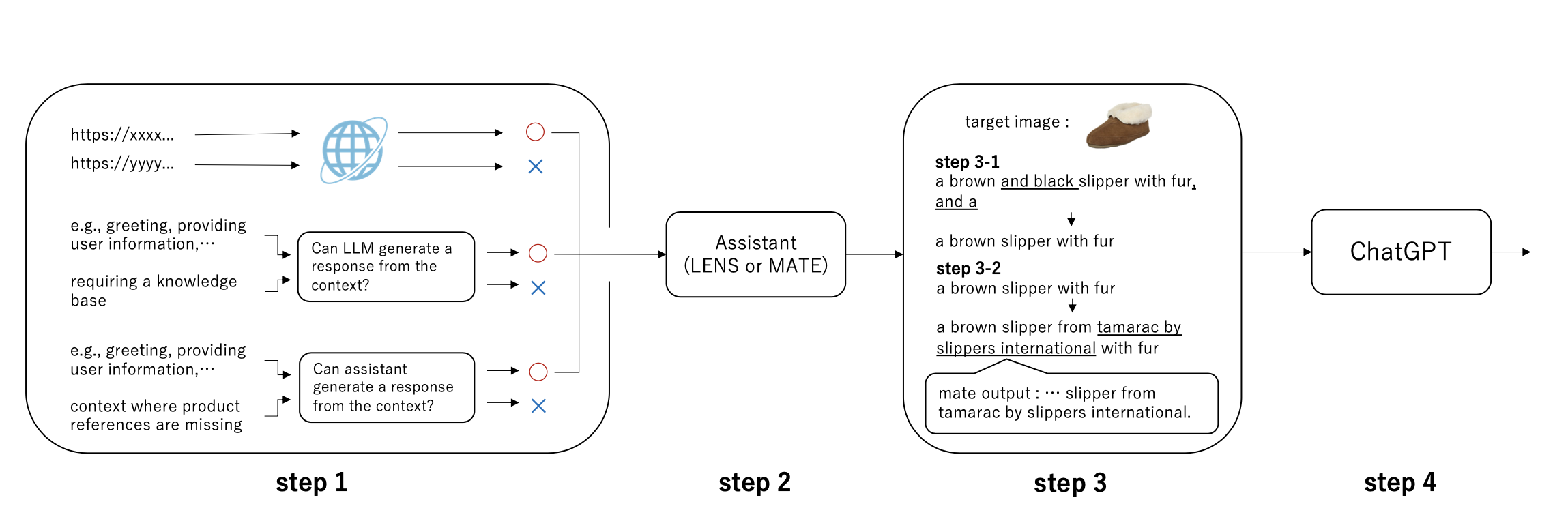}}
\caption{Experiment Steps}
\label{fig}
\end{figure*}

\subsection{Experiment Steps}
In the experiment, we conducted the following steps 1-4 (see Figure 4 for an overview) in order to enable multimodal task-oriented dialogues to be performed by each method, evaluation using G-EVAL, and fair experimentation and evaluation.
\begin{itemize}[left=5pt, labelwidth=*, align=left]
	\item[Step 1] Carefully select data where raw images are available, LLMs can generate a response from context, and each method can generate a response from context.
	\item[Step 2] Generate a response using each method.
	\item[Step 3] Modify image captions to be input to ChatGPT.
	\item[Step 4] Evaluate the response by using ChatGPT.
\end{itemize}
First, we explain the reason for selecting data under various conditions in step 1: The image data itself does not exist in the MMD dataset but links to images on the web are described (vectorized images by VGG \cite{b31} are included in the dataset). The method used in this study requires the use of raw images to generate captions, but meta information with broken links cannot be used, so the method is limited to images for which raw images can be used. In addition, this study does not use a knowledge base. Therefore, we excluded dialogues about anonymized celebrities and dialogues about sorting products, which are examples where evaluation would be meaningless if the knowledge base could not be used. Furthermore, we only selected data for which it is possible to generate utterances from the context, because there are examples where it is impossible to generate utterances from the context, i.e., images of products mentioned in the context are not present in the context. Step 3 is necessary because ChatGPT is used as the evaluator. First, it is currently not possible to input images into ChatGPT. Therefore, images are input as text in this study. The image captions were created based on BLIP2 but were found to be incomplete, so they were manually modified. The steps of the modification are as follows.
\begin{itemize}[left=5pt, labelwidth=*, align=left]
	\item[Step 3-1] Delete and correct grammatical errors and hallucination
	\item[Step 3-2] Add missing information in the caption
\end{itemize}
In step 3-1, grammatical errors and hallucinations are removed and corrected by looking at actual images and using search engines. Step 3-2 compares the response generated by MATE with the product information and images, and adds missing information to the captions.
The reason is that MATE is trained using training data from the MMD dataset and may have learned information about the product information and images that are not in the caption generated by BLIP2. If those information are not given as image captions, the evaluation is unfavorable to MATE. Therefore, we add those information to the captions so that a fair evaluation can be performed.

\begin{table}[htbp]
\caption{Experimental Results}
\begin{center}
\begin{tabular}{|c|c|c|c|c|}
\hline
& \textbf{\textit{Fluency}}& \textbf{\textit{Relevance}}& \textbf{\textit{Coherence}} & \textbf{\textit{Usefulness}}\\
\hline
\textbf{LENS}&\textbf{4.14}&\textbf{3.56}&\textbf{3.64}&\textbf{3.08}\\
\hline
\textbf{MATE}&3.60&3.30&3.38&2.64\\
\hline
\end{tabular}
\label{tab1}
\end{center}
\end{table}

\subsection{Experimental Results}
As shown in the results in Table 2, the LENS Framework significantly outperformed MATE, a previous Transformer-based model, on all four perspectives (fluency, relevance, coherence, and usefulness). Therefore, using LLMs appears to be more effective than learning from scratch with a large amount of dialogue data in this multimodal task-oriented dialogue scenario. Since we set up the evaluation system so that each item is judged as "good" if it exceeds 3 out of the 5 evaluation values from 1 to 5, the LENS Framework using an LLM is considered at least good on average in all the indicators.

\subsection{Discussion}
We believe that the quantity and quality of knowledge, the ability to understand images and language, and the ability to generate language in a pre-trained model trained on an even larger amount of data and with a larger parameter size than MATE led to the results of this experiment.\\
\indent Among the individual perspectives, fluency and usefulness have particularly high lifts. They have absolute lifts of 10.8\% and 8.8\%, respectively. A possible reason for the high fluency is that the fluency of the LLMs generating the utterances is very high (as we did not find any grammatical errors during our experiments). The 4.14 out of 5 rating, despite the very high fluency, may have been due to the number of tokens was limited for the simplicity of the experiment. The reason for the high usefulness is thought to be that usefulness requires high fluency, relevance, and coherence perspectives, as well as the quantity and quality of knowledge to provide useful information. Therefore, compared to previous studies, we believe that LENS has a higher lift, as each item is higher on average and it uses large-scale models. On the other hand, since usefulness is not high unless several perspectives are good, the absolute value of usefulness is small compared to the other perspectives.\\
\indent The lifts of relevance and coherence were not so large compared to the other two perspectives. We believe that this is due to the difficulty of dealing with the MMD dataset's unique dialogue (the assistant will state that it is going to search the catalog and provide a textual introduction of the product.), which is unlikely to have been learned by the pre-trained model. However, each has an absolute lift of 5.2\%.

\begin{figure}[htbp]
\centerline{\includegraphics[width=0.7\linewidth]{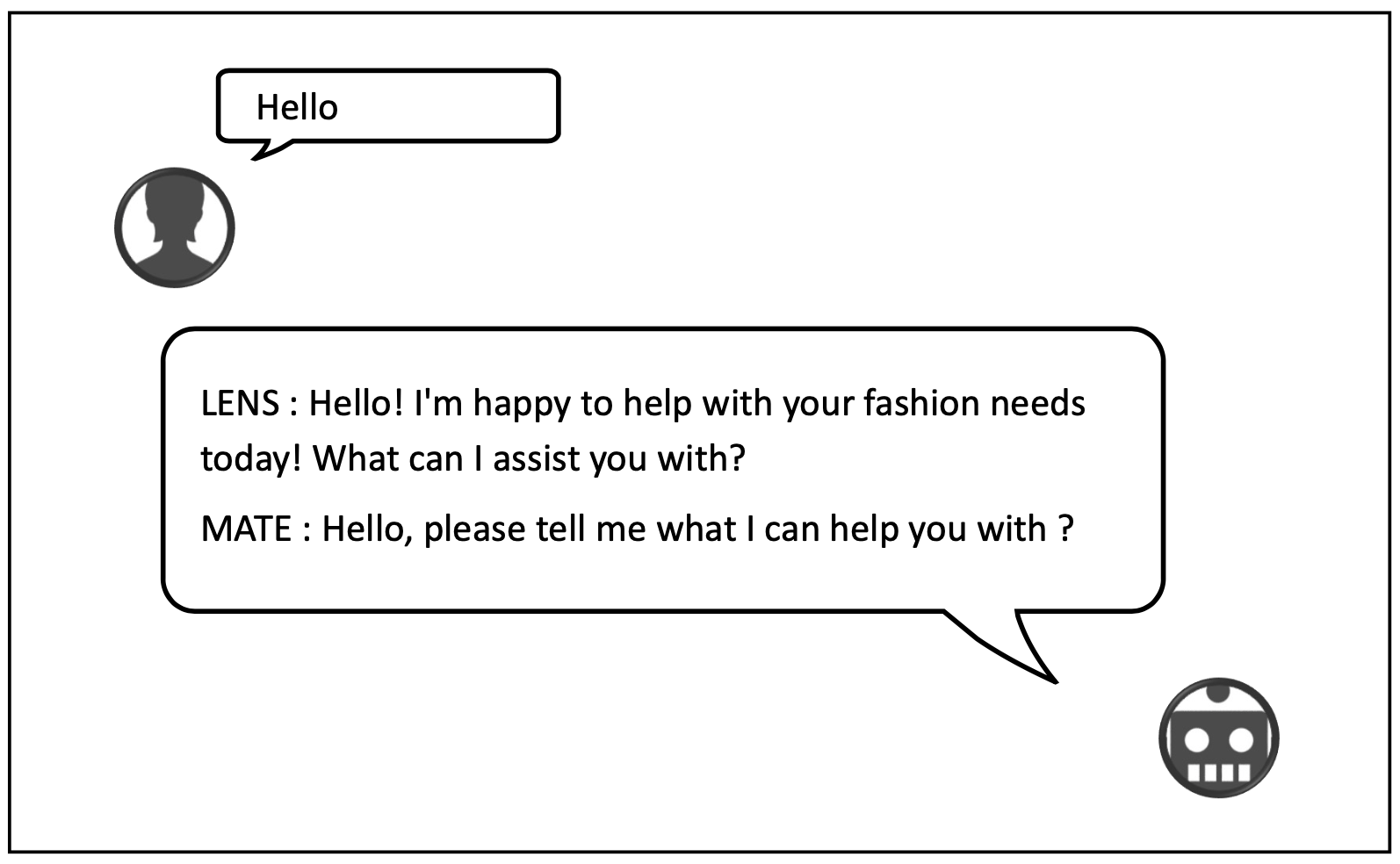}}
\caption{Example of general response}
\label{fig}
\end{figure}

\begin{figure}[htbp]
\centerline{\includegraphics[width=0.7\linewidth]{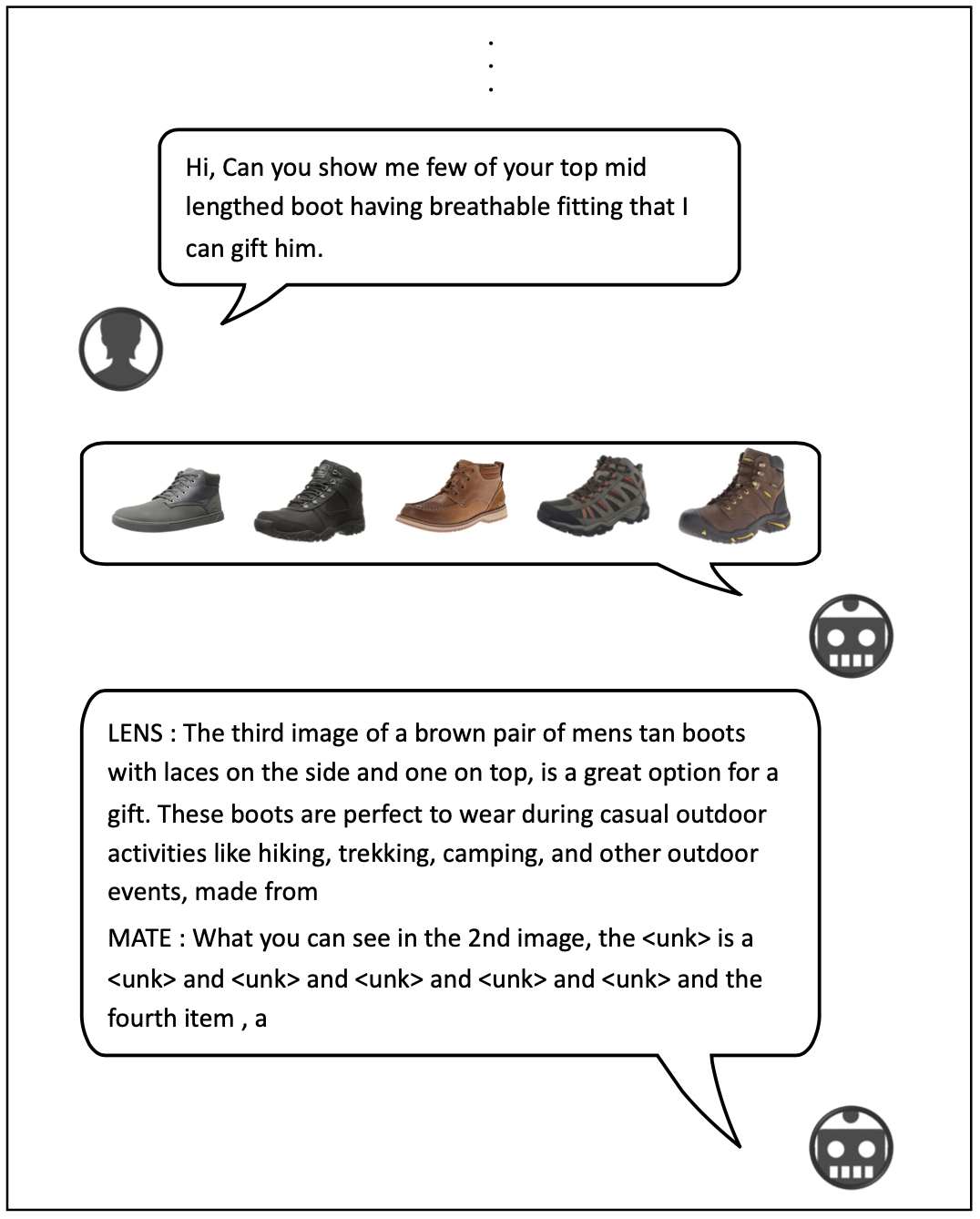}}
\caption{Example of successful LENS Framework response that take multimodal dialogue as the context}
\label{fig}
\end{figure}

\begin{figure}[htbp]
\centerline{\includegraphics[width=0.7\linewidth]{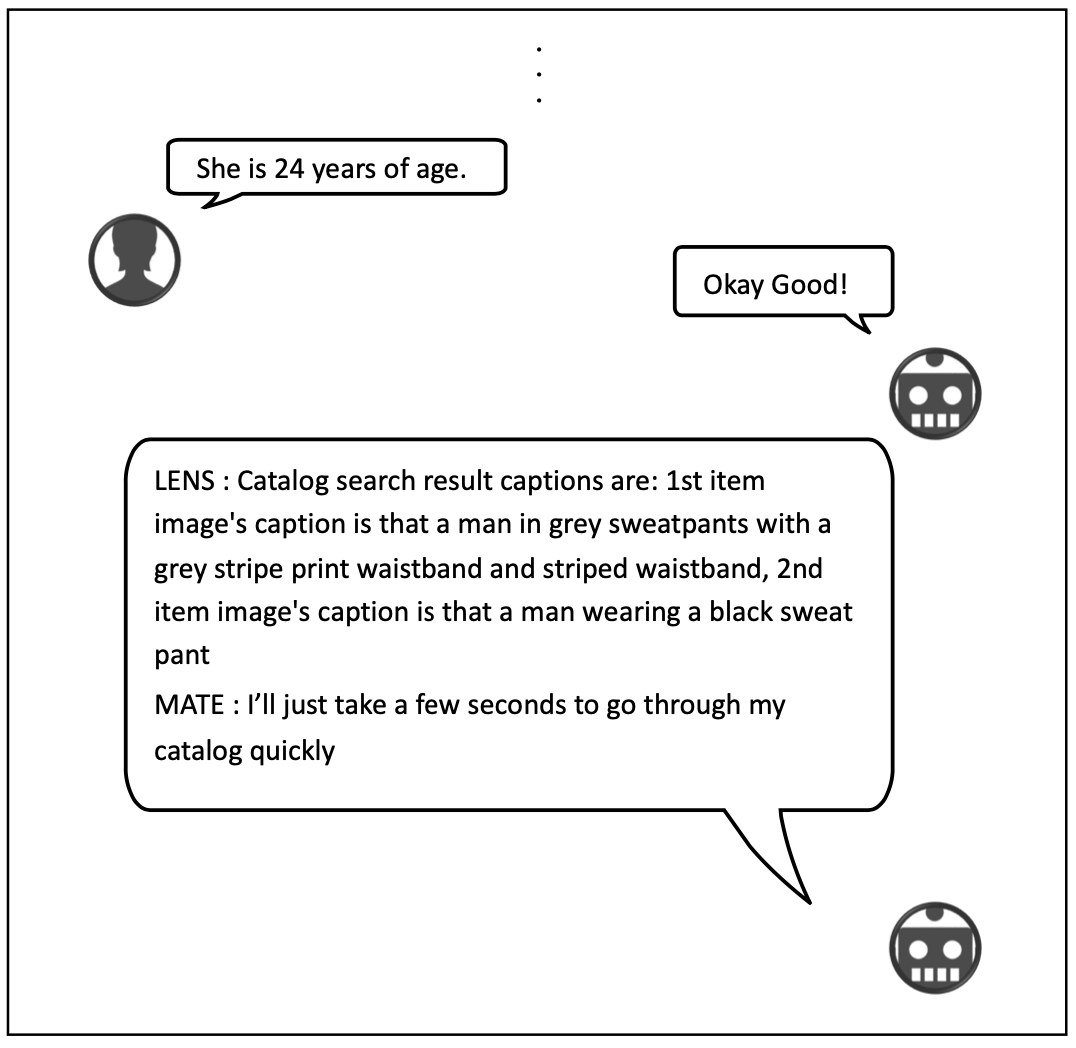}}
\caption{Example in which the LENS Framework failed}
\label{fig}
\end{figure}

\subsection{Case Study}
Figures 5-7 show examples of sampling from the test data. \\
\indent First, we analyze and discuss the general response (Figure 5), which takes only text as context and does not require much domain-specific knowledge. However, considering that the LENS Framework is less formal, and that the LENS Framework can be more useful depending on how the prompts are designed, we believe that LENS Framework is a step ahead of the other methods. \\
\indent Regarding the example of successful LENS Framework utterances that take multimodal dialogue as the context (Figure 6), the LENS Framework is able to use LLM's extensive knowledge to provide useful information such as "These boots are perfect to wear during casual outdoor activities like hiking, trekking, camping, and other outdoor events", although the sentences are broken due to the limited number of tokens. In contrast, MATE is somewhat fluent in grammar, but it outputs information that is completely useless. It should be added that not all of MATE's output is as incoherent as this example, and in many cases it produces useful information. \\
\indent Finally, we analyze and discuss the example in which the LENS Framework failed (Figure 7). The LENS Framework is commanded by the prompt to choose between two options, gather information about the user or declare that it will search the catalog. However, the LENS Framework ignored the instruction and even presented the results of a fictitious catalog search. One possible reason for not following the instruction is that the action of declaring to search the catalog itself is specific to the MMD dataset. In contrast, MATE has been trained on the MMD dataset, so it is clear that it is able to respond appropriately.

\subsection{Limitations}
The first drawback of the LENS Framework constructed in this study is hallucination. Even though we devised a system that does not cause hallucination as a prompt, as shown in the previous subsection, we found examples of outputting information such as fictitious sizes of products and fictitious materials.\\
\indent Second, the LLM used in the LENS Framework did not follow the instructions. In situations where we wanted the LLM to take the desired action, as shown in the previous subsection, there were instances where they did not follow the instructions and took their own actions. This may be due to the fact that the examples were given in a fixed one-shot format, the number of examples was small to begin with, the context was only given for the two previous turns of dialogue, and the LLM used (parameter size and training data). \\
\indent In this study, the previous two turns of dialogue were used to generate utterances, which is impractical in that it does not use all contexts. We plan to conduct further experiments and evaluations in this regard.\\
\indent The method of using ChatGPT as an evaluator has also just begun to be studied; it has been reported \cite{b25} that people tend to evaluate the output of LLMs more highly than human-written text, and we believe that further investigation is needed.\\
\indent Furthermore, there is the problem that the evaluation using the dataset itself is not exactly the same as the real-world application. In this study, we considered the MMD dataset as a real-world scenario and evaluated the LENS Framework. However, it is difficult to fully evaluate the performance of LLMs with real-world applications in mind using this method. Therefore, a new method of evaluation will be needed.

\section{Conclusion}
In this study, we tested the potential of the LENS Framework, which uses pre-trained models with fixed parameters to perform multimodal tasks, against MMD, a benchmark for multimodal task-oriented dialogues. Experimental results showed that the LENS Framework outperformed its Transformer-based predecessor, MATE, on all four perspectives: fluency, relevance, coherence, and usefulness. In particular, significant lifts were observed in fluency and usefulness. This may be due to the high fluency of the LLMs, their ability to achieve high accuracy on a variety of perspectives other than fluency, and the quantity and quality of knowledge from large-scale models. \\
\indent However, problems with hallucination and failure to follow instructions also arose with the LENS Framework. Further research is also needed to determine whether ChatGPT should be used as an evaluator and how it should be evaluated. \\
\indent In the future, we intend to conduct experiments using all past contexts, use a knowledge base, explore new evaluation methods, and validate the use of various models.

\section*{Appendix}
\begin{figure*}[htbp]
\centerline{\includegraphics[width=0.8\linewidth]{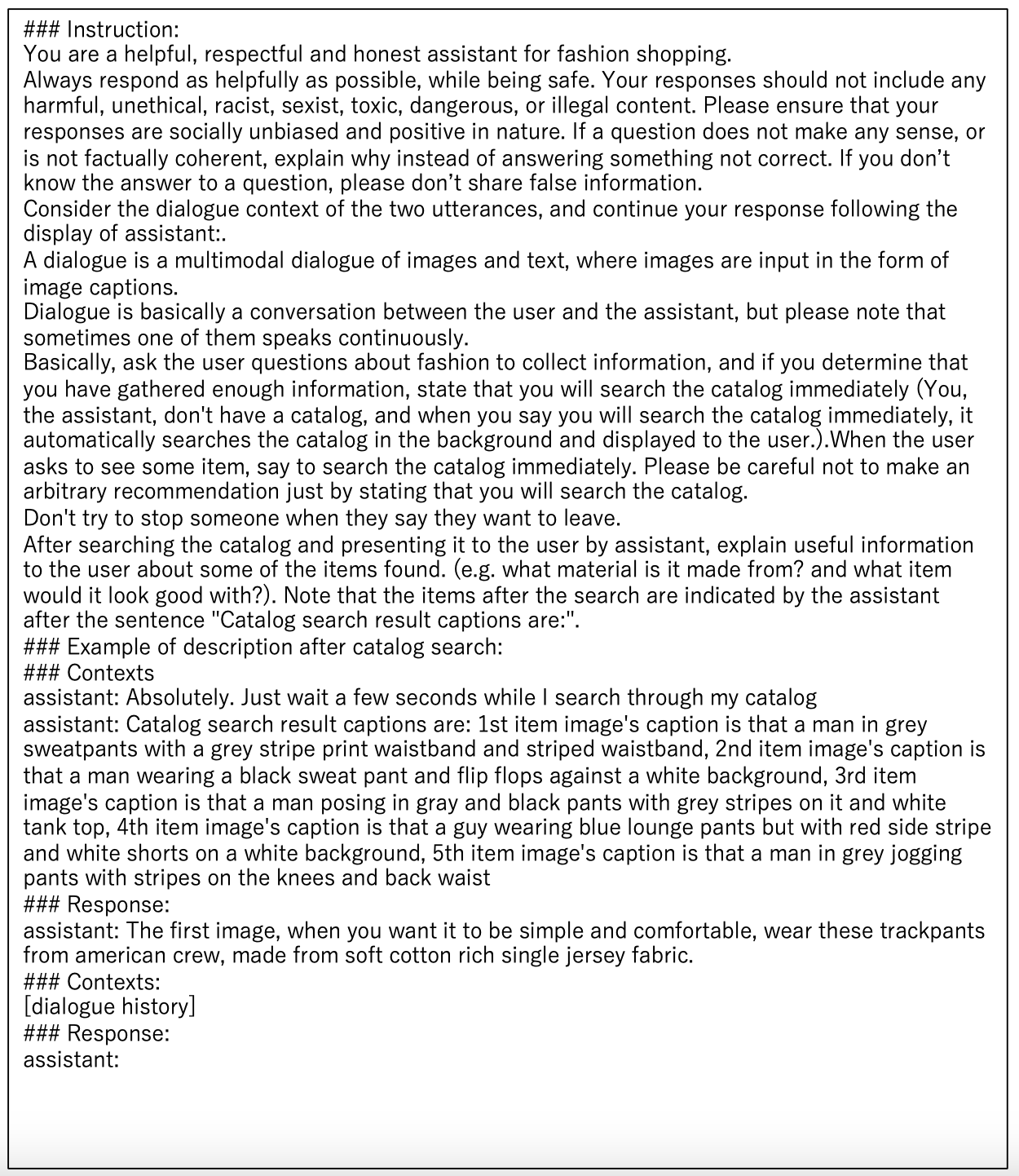}}
\caption{Prompt (LENS Framework)}
\label{fig}
\end{figure*}

\begin{figure*}[htbp]
\centerline{\includegraphics[width=0.8\linewidth]{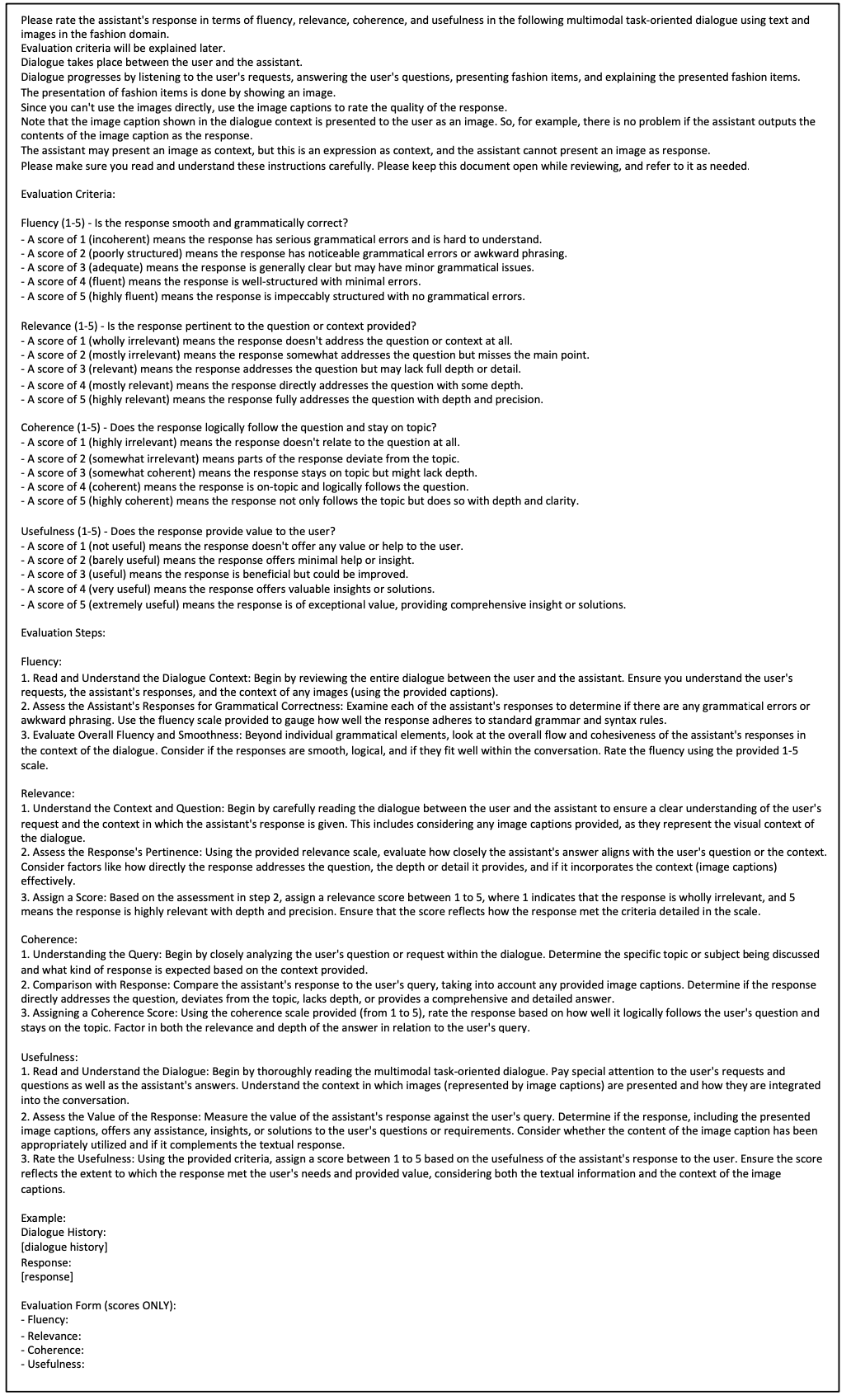}}
\caption{Prompt (G-EVAL)}
\label{fig}
\end{figure*}

\end{document}